\documentclass{article} 
\usepackage{iclr2020_conference,times}

\usepackage{amsmath,amsfonts,bm}

\def\eqref#1{equation~\ref{#1}}

\def\1{\bm{1}}

\DeclareMathAlphabet{\mathsfit}{\encodingdefault}{\sfdefault}{m}{sl}
\SetMathAlphabet{\mathsfit}{bold}{\encodingdefault}{\sfdefault}{bx}{n}

\usepackage{hyperref}
\usepackage{url}
\usepackage{graphicx}
\usepackage{amsmath}
\usepackage{booktabs}
\usepackage{xcolor}

\usepackage[symbol]{footmisc}

\title{Binarized Neural Networks for Resource-Constrained On-Device Gait Identification}

\author{Daniel J. Wu \thanks{Equal contribution}\\
Stanford Computer Science\\
\texttt{danjwu@stanford.edu} \\
\And
Avoy Datta \footnotemark[1] \\
Stanford Electrical Engineering \\
\texttt{avoy.datta@stanford.edu} \\
\And
Vinay Prabhu \\
UnifyID Co. \\
\texttt{vinay@unify.id}
}

\iclrfinalcopy 
\begin{document}

\maketitle
\begin{abstract}
User authentication through gait analysis is a promising application of discriminative neural networks -- particularly due to the ubiquity of the primary sources of gait accelerometry, in-pocket cellphones. However, conventional machine learning models are often too large and computationally expensive to enable inference on low-resource mobile devices. We propose that binarized neural networks can act as robust discriminators, maintaining both an acceptable level of accuracy while also dramatically decreasing memory requirements, thereby enabling on-device inference. To this end, we propose BiPedalNet, a compact CNN that nearly matches the state-of-the-art on the Padova gait dataset, with only \textbf{1/32} of the memory overhead.

\end{abstract}

\section{Introduction}

\subsection{Gait Identification}

Human gait, as measured via smartphone sensors, has been shown to be a viable biometric for distinguishing users with high accuracy (\cite{derawi2010unobtrusive}). Both user privacy and scalability concerns suggest re-identification is best done on-device. However, previous work on user re-identification from gait data consistently demonstrates high accuracy with deep CNNs, but also high computational overhead, making on-device, real-time, inference, intractable (\cite{IDNet}).

\subsection{Binarized Neural Networks}
Binarized Neural Networks (\cite{BNNs}) are neural networks with weights constrained to $\{-1, 1\}$. BNNs, although generally achieving lower accuracies than their full-precision equivalents, are both smaller, due to compact binary weight matrices, and faster, due to the usage of bitwise operations for matrix multiplication and activation functions. 

This makes BNNs an ideal choice for on-device machine learning. Mobile devices, particularly budget devices, often have strict limitations on both memory availability and processing power, making frugality an important virtue. We propose binarized neural networks as a computationally inexpensive way to conduct lightweight gait identification.

\section{Methodology}
\vspace{-0.1in}
\subsection{Training Binarized Weights}
In order to train the binarized weights of the architecture, we maintain latent real-valued weights proposed in \cite{BNNs} that are updated during backpropagation and binarized during the forward pass. These latent weights are set to 1 if positive; otherwise, clipped to -1 in the forward pass. Note that these latent weights are not used at test time, and thus don't contribute to the memory overhead. 

\subsection{Architecture}

We propose a lightweight, binarized neural network architecture, dubbed BiPedalNet. BiPedalNet is built primarily using convolutional neural layers. Having 2-dimensional convolutions early in the network allows the model to learn cross-channel dependencies between the accelerometric and gyroscopic dimensions. Once the cross-channel features are extracted, we restrict the convolutions to a single dimension to confine learning to the temporal dimension. For regularization and pooling, we found the combination of max pooling and batch normalization yields the best performance. The binarized architecture is implemented with the LARQ (\cite{larq}) library.

  
\begin{table}
    \centering
    \includegraphics[height=1.5in]{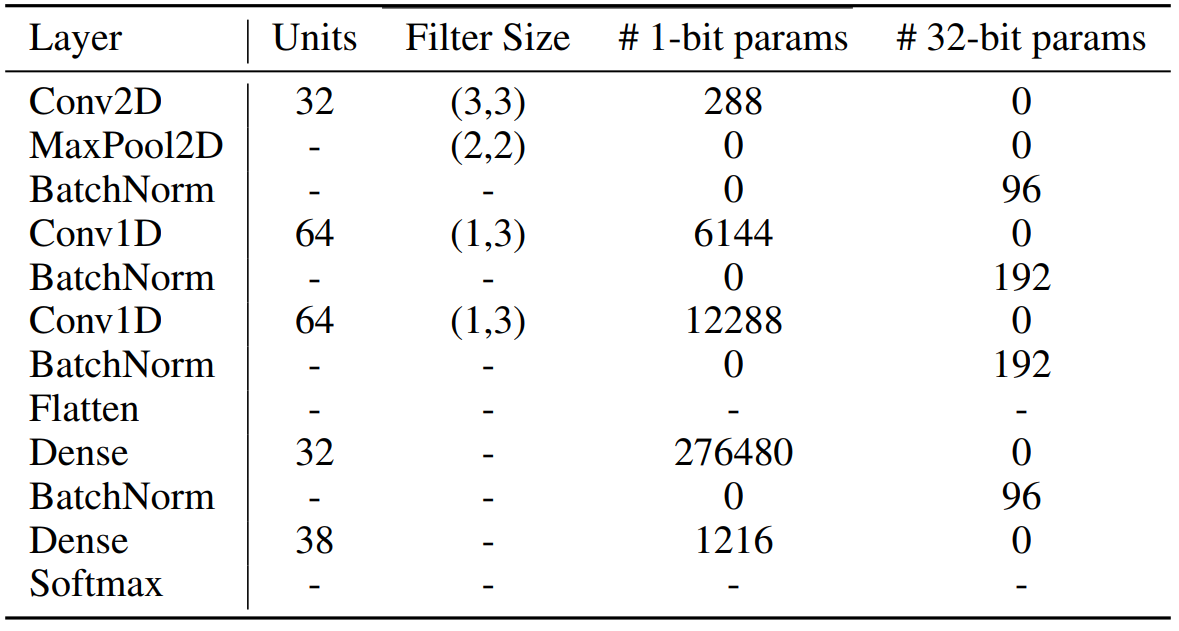}
    \caption{The BiPedalNet Architecture. All layers have binary weights.}
    \label{tab:BiPedalNet}
\end{table}

\subsection{Training Procedure}

To compare our architecture against the existing state-of-the-art, we train and validate our architecture on the Padova gait dataset \cite{IDNet} (80/20 split). We compare BiPedalNet against a binzarized version of the IDNet model proposed in \cite{IDNet}, with both models trained using the same hardware (1x Tesla V100 GPU). 

We use the Adam optimizer (\cite{adam}) to update the latent weights of the architecture, with a default learning rate of 1e-3 and a scheduler that exponentially reduces the rate when validation accuracy plateaus. The multi-class cross-entropy metric is used as the optimization objective.

\section{Results and Discussion}
\vspace{-0.1in}
\begin{table*}[!htb]
 \centering
  \begin{tabular}{lccc}
    \toprule
     \cmidrule(r){3-4}
     Model  & Number of Parameters & Size on disk (MB) & Top-1 Accuracy(\%) \\
    \midrule
        IDNet & 335k & 1.28 & 98.05\\
        Binarized IDNet & 335k & 0.04 & 41.45\\
        BiPedalNet & 297k& 0.04 & \textbf{95.91}\\    
    \bottomrule
    \end{tabular}
  \caption{Quantitative Results for performance on Padova dataset}
   \label{table:results}
  \end{table*}
  
The first two results in Table \ref{table:results} demonstrate that state-of-the-art performance on gait data does not persist through off-the-shelf binarization of full-precision networks. To achieve performance comparable to the full-precision model, special care needs to be applied in tuning the architecture to both the dataset and the task at hand. This is demonstrated by the superior performance of BiPedalNet over the binarized IDNet variant.

\section{Conclusion}
We've shown that binarized neural networks can achieve comparable accuracies to full precision architectures, at a fraction of the size, on gait identification tasks. However, it is clear that custom-designing binarized architectures is necessary for reasonable performance -- naively binarizing a given full precision architecture is a recipe for disaster. Binarized neural networks offer a promising avenue to achieve low latency, low memory, inference in hardware-constrained devices, and we expect these models to perform well in low-resource mobile devices.

\newpage
\bibliography{egbib}
\bibliographystyle{iclr2020_conference}

\end{document}